# The Linguistic DS: Linguisitic Description in MPEG-7

**HASIDA, Kôiti**

**CARC, AIST and CREST, JST**

Contents

## 1 Introduction

MPEG-7 (Moving Picture Experts Group Phase 7) is an XML-based international standard on semantic description of multimedia content. This document discusses the Linguistic DS and related tools. The linguistic DS is a tool, based on the GDA tag set (http://i-content.org/GDA/tagset.html), for semantic annotation of linguistic data in or associated with multimedia content. The following text reflects `Study of FPDAM – MPEG-7 MDS Extensions' issued in March 2003, and not most part of MPEG-7 MDS, for which the readers are referred to the first version of MPEG-7 MDS document available from ISO (http://www.iso.org). Without that reference, however, this document should be mostly intelligible to those who are familiar with XML and linguistic theories. Comments are welcome and will be considered in the standardization process.

## 2 Multimedia content entity description tools

Specified here is how to embed linguistic descriptions in a MPEG-7 file.

### 2.1 Multimedia content entity description tools syntax

```
<!-- ####################################################### -->
<!--  Definition of Multimedia content entity tools (4.4.5)   -->
<!-- ####################################################### -->
<!-- Definition of Linguistic Content Entity (AMD/1) -->
<complexType name="LinguisticType">
  <complexContent>
    <extension base="mpeg7:MultimediaContentType">
      <sequence>
        <element name="Linguistic" type="mpeg7:LinguisticDocumentType"/>
      </sequence>
    </extension>
  </complexContent>
</complexType>
```

### 2.2 Multimedia content entity description tools semantics

Semantics of the `LinguisticType`:

| *Name* | *Definition* |
|---|---|
| `LinguisticType` | Multimedia content entity for describing linguistic content. Linguistic content refers to textual, spoken, visually written, and other information that is modelled from the natural language point of view. `LinguisticType` extends `MultimediaContentType`. |
| `Linguistic` | Describes the linguistic content. The linguistic content may correspond to a full linguistic document or a segment. `Linguistic` is of type `LinguisticDocumentType`. |

### 2.3 Multimedia content entity description tools examples (informative)

The following example shows the use of the multimedia content entity `LinguisticType` for describing linguistic content. The example is a whole MPEG-7 file containing a linguistic annotation of the sentence: “Tom visited his mother.”

```
<Mpeg7>
  <!-- MDS AMD/1 Example -->
  <Description xsi:type="ContentEntityType">
    <MultimediaContent xsi:type="LinguisticType">
      <Linguistic>
        <Paragraph>
          <Sentence>
            <Phrase id="Tom">Tom </Phrase>
            visited
            <Phrase><Phrase equal="#Tom">his </Phrase>mother </Phrase>.
          </Sentence>
        </Paragraph>
      </Linguistic>
    </MultimediaContent>
  </Description>
</Mpeg7>
```

# 3 GeneralRelation DS

## 3.1 Introduction

The `GeneralRelation DS` extends the `Relation DS` (in the first version of MPEG-7 MDS) by providing a `typelist` attribute to accommodate multiple relation terms and `generalSource` and `generalTarget` attributes to accommodate `termReferenceType` arguments.

## 3.2 GeneralRelation DS syntax

```
<!-- ##################################################### -->
<!-- Definition of GeneralRelation DS (7.7)                -->
<!-- ##################################################### -->

<complexType name="GeneralRelationType">
  <complexContent>
    <extension base="mpeg7:RelationType">
      <attribute name="typelist" type="mpeg7:termReferenceListType" use="optional"/>
      <attribute name="generalSource"
         type="mpeg7:termReferenceListType" use="optional"/>
      <attribute name="generalTarget"
         type="mpeg7:termReferenceListType" use="optional"/>
    </extension>
  </complexContent>
</complexType>

<simpleType name="termReferenceListType">
   <list itemType="mpeg7:termReferenceType"/>
</simpleType>
```

## 3.3 GeneralRelation DS semantics

| *Name* | *Definition* |
|---|---|
| `GeneralRelationType` | Describes a generalized relation among DS instances and controlled terms. The arguments of a relation are categorized into the `generalSource` (first) and the `generalTarget` (second) arguments. Both the `generalSource` and `generalTarget` arguments for a relation are ordered. A relation may have a variable number of arguments in each of the source and the target. |
| `typelist` | Identifies the relation type by a list of controlled terms in some classification schemes. The default interpretation of this list is the sequential composition of the binary relations represented by the controlled terms. Non-default interpretations should be stipulated in the definitions of individual controlled terms. |
| `generalSource` | Refers to the DS instances and/or controlled terms that are the source argument(s) of the relation. The elements contained in this attribute are ordered so that the first one identifies the first source argument, the second one the second argument, and so on. |
| `generalTarget` | Refers to the DS instances and/or controlled terms that are the target argument(s) of the relation. The items contained in this attribute are ordered so that the first one identifies the first target argument, the second one the second argument, and so on. |
| `termReferenceListType` | Specifies a sequence of controlled-term references. |

## 3.4 GeneralRelation DS examples (informative)

The `Relation` elements in the examples below are assumed to be of `GeneralRelationType`.

```
<Relation typelist="urn:mpeg:mpeg7:cs:SemanticRelationCS:2001:source
```

```
              urn:mpeg:mpeg7:cs:SpatialRelationCS:2001:below"
        source="#anAppearanceEvent" target="#aTable"/>
```

This reads that something appears from below a table (what appears is not explicitly referred to in this description). That is, the `typelist` attribute defaults to represent the composition of the binary relations represented by the controlled terms in its value, unless otherwise specified in the definition of the individual controlled terms. Namely, `typelist=`"$R_1$ $R_2$ ... $R_n$" by default means that there are entities $A_0$, ..., $A_n$ such that $A_{i-1}$ $R_i$ $A_i$ ($A_{i-1}$ and $A_i$ as the first and the second arguments, respectively, are in relation $R_i$) for $0 < i <= n$, $A_0$ is the source argument, and $A_n$ is the target argument of the current relation.

This extension raises the efficiency of descriptions based on the Linguistic DS, the Semantic DS, and so forth, in the sense that we can skip explicit descriptions of the intermediate entities ($A_2$, ..., $A_{n-1}$) in the composition; in the above example, we avoid explicit description of the location below the table, which is the source location of the appearance event.

A non-default example follows, which means that Tom is the first of those who gather.

```
<Relation typelist="urn:SomeOntologyOfRelations:object
                    urn:SomeOntologyOfRelations:initial"
        source="#aGatheringEvent" target="#Tom"/>
```

Here, controlled term `initial` is not simply composed with `object`, but interpreted relative to both the object (the gathering people) and the gathering event. That is, the initiality is based on some ordering in the object of the event and this ordering is induced by the event. This should be stipulated in the definition of `initial`.

Below is another non-default example.

```
<Relation typelist=":r:r1 :meta:and :r:r2" source="#s" target="#t"/>
```

If controlled term `and` is stipulated to form the logical conjunction (or the intersection in the set-theoretic terminology) of the binary relations on both sides of it, then the above description means the logical conjunction of `s r1 t` (`s` and `t` stand in relation `r1`) and `s r2 t` (`s` and `t` stand in relation `r2`).

The `generalSource` and `generalTarget` attributes are generalized versions of the `source` and `target` attributes in Relation DS, respectively. The values of the `generalSource` and `generalTarget` attributes are lists of `anyURI` elements, whereas the values of the `Source` and `generalTarget` attributes are lists of `termReferenceType` elements; Note that `termReferenceType` subsumes `anyURI`. The `generalSource` and `generalTarget` attributes can hence reference controlled terms, which may represent not only concepts but also individuals. For instance, the following example means that the event represented by `event1` took place in New York

```
<Relation type="urn:mpeg:mpeg7:cs:SemanticRelationCS:2001:location"
   source="#event1" generalTarget="urn:SomeOntologyOfPlaces:NewYork"/>
```

Another example, shown below, involves a deixis (a reference determined by the local context; natural-language examples are *here*, *now*, *me*, *you*, *today*, *last year*, and so on). The target of this relation is the previous sibling element *X* of the element *Y* embedding this `Relation` element, which means that what *Y* represents occurs after what *X* represents.

```
<Relation type="urn:mpeg:mpeg7:cs:TemporalRelationCS:2001:after"
   generalTarget="urn:SomeOntologyOfDeixes:previous"/>
```

# 4 Linguistic description

## 4.1 Introduction

This subclause specifies a tool for describing linguistic content.

The `Linguistic DS` describes the semantic structure (in particular, propositional content and dialogue structure) of linguistic data associated with multimedia content. Linguistic data can take the form of textual, written or spoken content and may represent transcriptions, critiques, scenarios, and so forth. Some multimedia content may consist of textual linguistic data as the main component and additional video and/or audio data. So the `Linguistic DS` is designed to address linguistic data in general.

The `Linguistic DS` hence provides greater descriptive power than Textual Annotation tools defined in ISO/IEC 15938-5. In particular, the `Linguistic DS` is an upward compatible extension of the `DependencyStructure datatype` (specified in the first version of MPEG-7 MDS) in the sense that every instance of the `DependencyStructure datatype` can be mapped to an equivalent instance of the `Linguistic DS`. In addition, the `Linguistic DS` extends the functionality of the `DependencyStructure datatype` in several respects. The `Linguistic DS` can describe linguistic entities larger than sentences, such as paragraphs and chapters. This is necessary for dealing with the structure of an entire document. Although the `DependencyStructure datatype` addresses dependencies inside sentences and the notion of such dependency is basically shared between `DependencyStructure datatype` and the `Linguistic DS`, the `Linguistic DS` also addresses dependencies outside of sentences: i.e., among linguistic entities such as sentences, groups of sentences, paragraphs, sections, chapters, and so on. When a sentence represents the cause or reason of the event represented by another sentence, for instance, the former sentence may be regarded as depending on the latter.

There are many aspects of linguistic descriptions such as phonetics, phonology, morphology, syntax, semantics, and pragmatics. However, the Linguistic DS focuses on the syntactic and the semantic aspects only, in order to interpret propositional content and dialogue structure, because they are most essential for the sake of practical content services such as retrieval, summarization, presentation, and so on. Nevertheless, the `Linguistic DS` provides a general container of various aspects of linguistic descriptions other than syntax and semantics. The scope of linguistic descriptions can be expanded by introducing new ontologies or classification schemes. For example, consider the following dialog consisting of a question and a reply. The `Linguistic DS` can be used to describe the dialogue relation between the sentences by identifying that the reply sentence is related to question sentence through a “reply” relationship, where this “reply” relation is defined in a new classification scheme for dialog relations.

```
<Mpeg7>
   <!-- MDS AMD/1 Example -->
   <Description xsi:type="ContentEntityType">
      <MultimediaContent xsi:type="LinguisticType">
         <Linguistic>
            <Paragraph>
               <Quotation>
                  <Sentence id="Q">Are you ready? </Sentence>
               </Quotation>
               <Quotation>
                  <Sentence>
                     <Relation type="urn:DialogueRelations:reply" target="#Q"/>
                     Yes.
                  </Sentence>
               </Quotation>
            </Paragraph>
         </Linguistic>
      </MultimediaContent>
   </Description>
</Mpeg7>
```

Descriptions based on the `Linguistic DS` do not require any linguistic data at all. The `Linguistic DS` may therefore be used in place of the `Semantic DS`. For instance, the following description means that a person called Hasida Koiti sleeps. (ISO 8601, an international standard on the representation of dates and times, is a vocabulary of infinitely many terms and hence cannot be defined as a ClassificationScheme in ISO/IEC 15938-5 (the first version of MPEG-7 MDS), but can be referenced as below if properly registered and thus publicly resolvable.)

```
<Mpeg7>
  <Description xsi:type="ContentEntityType">
    <MultimediaContent xsi:type="LinguisticType">
      <Linguistic>
        <Sentence semantics="urn:SomeOntologyOfEvents:sleep">
          <Relation type="urn:mpeg:mpeg7:cs:SemanticRelationCS:2001:time"
                    target="urn:ISO8601:2003-02-13"/>
          <Phrase semantics="urn:SomeOntologyOfObjects:person">
```

```
          <Relation type="urn:SomeOntologyOfAttributes:familyName"
                    target="urn:OntologyOfASCIItexts:Hasida"/>
          <Relation type="urn:SomeOntologyOfAttributes:givenName"
                    target="urn:OntologyOfASCIItexts:Koiti"/>
        </Phrase>
      </Sentence>
    </Linguistic>
  </MultimediaContent>
 </Description>
</Mpeg7>
```

Furthermore, the `copy` and the `substitute` attributes introduced below straightfowardly address abstraction and instantiation. In this sense, the Linguistic DS is a concise tool for describing any semantic content, whether or not verbally expressed.

## 4.2 Linguistic DS syntax

```
<!-- ##################################################### -->
<!-- Definition of Linguistic DS (7.11.2) (AMD/1)          -->
<!-- ##################################################### -->
<complexType name="LinguisticEntityType" abstract="true" mixed="true">
   <complexContent mixed="true">
      <extension base="mpeg7:DSType">
         <sequence>
            <element name="MediaLocator" type="mpeg7:MediaLocatorType"
               minOccurs="0" maxOccurs="unbounded"/>
            <element name="Relation" type="mpeg7:GeneralRelationType"
               minOccurs="0" maxOccurs="unbounded"/>
         </sequence>
         <attribute ref="xml:lang"/>
         <attribute name="start" use="optional">
            <simpleType>
               <union memberTypes="nonNegativeInteger mpeg7:mediaTimePointType"/>
            </simpleType>
         </attribute>
         <attribute name="length" use="optional">
            <simpleType>
               <union memberTypes="nonNegativeInteger mpeg7:mediaDurationType"/>
            </simpleType>
         </attribute>
         <attribute name="type" type="mpeg7:termReferenceType" use="optional"/>
         <attribute name="depend" type="mpeg7:termReferenceType" use="optional"/>
         <attribute name="equal" type="mpeg7:termReferenceListType" use="optional"/>
         <attribute name="semantics" type="mpeg7:termReferenceListType"
            use="optional"/>
         <attribute name="compoundSemantics" type="mpeg7:termReferenceListType"
            use="optional"/>
         <attribute name="operator" type="mpeg7:termReferenceListType"
            use="optional"/>
         <attribute name="copy" type="mpeg7:termReferenceListType" use="optional"/>
         <attribute name="noCopy" type="mpeg7:termReferenceListType"
            use="optional"/>
         <attribute name="substitute" type="mpeg7:termReferenceType"
            use="optional"/>
         <attribute name="inScope" type="mpeg7:termReferenceType" use="optional"/>
         <attribute name="edit" use="optional">
            <simpleType>
               <restriction base="NMTOKEN">
                  <whiteSpace value="preserve"/>
                  <pattern value=":.*"/>
               </restriction>
            </simpleType>
```

```
          </attribute>
       </extension>
   </complexContent>
</complexType>
<!-- ##################################################-->
<!-- Definition of LinguisticDocument DS                 -->
<!-- ##################################################-->
<complexType name="LinguisticDocumentType">
   <complexContent>
      <extension base="mpeg7:LinguisticEntityType">
         <choice maxOccurs="unbounded">
            <element name="Heading" type="mpeg7:SentencesType"/>
            <element name="Division" type="mpeg7:LinguisticDocumentType"/>
            <element name="Paragraph" type="mpeg7:SentencesType"/>
            <element name="Sentences" type="mpeg7:SentencesType"/>
            <element name="Sentence" type="mpeg7:SyntacticConstituentType"/>
            <element name="Quotation" type="mpeg7:LinguisticDocumentType"/>
         </choice>
         <attribute name="synthesis" type="mpeg7:synthesisType" use="optional"
            default="coordination"/>
      </extension>
   </complexContent>
</complexType>
<!-- ##################################################-->
<!-- Definition of Sentences DS                          -->
<!-- ##################################################-->
<complexType name="SentencesType">
   <complexContent>
      <extension base="mpeg7:LinguisticEntityType">
         <choice minOccurs="0" maxOccurs="unbounded">
            <element name="Sentences" type="mpeg7:SentencesType"/>
            <element name="Sentence" type="mpeg7:SyntacticConstituentType"/>
            <element name="Quotation" type="mpeg7:LinguisticDocumentType"/>
         </choice>
         <attribute name="synthesis" type="mpeg7:synthesisType" use="optional"
            default="coordination"/>
      </extension>
   </complexContent>
</complexType>
<!-- ##################################################-->
<!-- Definition of SyntacticConsituent DS                -->
<!-- ##################################################-->
<complexType name="SyntacticConstituentType">
   <complexContent>
      <extension base="mpeg7:LinguisticEntityType">
         <choice minOccurs="0" maxOccurs="unbounded">
            <element name="Head" type="mpeg7:SyntacticConstituentType"/>
            <element name="Phrase" type="mpeg7:SyntacticConstituentType"/>
            <element name="Quotation" type="mpeg7:LinguisticDocumentType"/>
         </choice>
         <attribute name="baseForm" type="string" use="optional"/>
         <attribute name="synthesis" type="mpeg7:synthesisType" use="optional"
            default="dependency"/>
         <attribute name="functionWord" type="string" use="optional"/>
      </extension>
   </complexContent>
</complexType>
<!-- ##################################################-->
<!-- Definition of Synthesis DataType                    -->
<!-- ##################################################-->
<simpleType name="synthesisType">
   <restriction base="string">
      <enumeration value="none"/>
      <enumeration value="dependency"/>
```

```
      <enumeration value="forward"/>
      <enumeration value="backward"/>
      <enumeration value="coordination"/>
      <enumeration value="apposition"/>
      <enumeration value="repair"/>
      <enumeration value="error"/>
   </restriction>
</simpleType>
```

## 4.3 Linguistic DS semantics

Semantics of the Linguistic DS follows. Instances of LinguisticEntityType are called **Linguistic elements**.

| Name | Definition |
|---|---|
| LinguisticEntityType | Describes linguistic content (abstract). The specialized linguistic description tools extend the LinguisticEntity DS. In the typical usage, LinguisticEntityType mixes the description and the text representation of the language data being described, "marking up" the text with tags. LinguisticEntityType extends DSType. |
| MediaLocator | Describes a locator of the linguistic content (optional). The MediaLocator can be used to locate different forms of linguistic content such as speech, text, or visual text data and indicate the corresponding location in time or stream. |
| Relation | Describes a relation of the parent element with the target argument specified by the target or generalTarget attribute when the source and generalSource attributes are absent (optional). The target argument may be an instance (or a set of instances) of the LinguisticEntity DS, the Segment DS, and the Semantic DS.<br><br>The type and typelist attributes partially specify the relation in question, in the sense that the exact relation is the composition *prq* of a prefix relation *p*, a postfix relation *q*, and the relation *r* specified by the attribute. *A p B* (*B q A*) means either *A* is equal to *B*, *B* is an instance, member, part or subset of *A*, or *B* is referred to by referring to *A* (metonymy). |
| xml:lang | Indicates the language of the linguistic data being described (optional). If no value is specified for xml:lang. Its value is inherited from the minimal ancestor element that specifies a value for xml:lang. |
| start | Indicates the start point of the current linguistic entity in the file specified by the nearest preceding MediaLocator element which is a child of an ancestor to the current element. When a start attribute has a nonNegativeInteger value, its unit is the one used most recently in the above-mentioned MediaLocator element. |
| length | Specifies the length or duration of the current linguistic entity. When a length attribute has a nonNegativeInteger value, its unit is the one used most recently in the nearest preceding MediaLocator element which is a child of an ancestor to the current element |
| type | Indicates the type of the linguistic entity (optional). Examples of type include document parts such as chapter, section, paragraph, or parts of speech such as noun, verb, and so forth. |

| Name | Definition |
| --- | --- |
| `depend` | Indicates the dependency of the current linguistic entity on another linguistic entity (optional). Namely, indicates that the governor node (see 4.1) of the current element is equal to the self node (see 4.1) of the referenced Linguistic element. The `depend` attribute is attached to an extraposed linguistic entity depending on another entity embedded in one of its siblings.<br><br>Sentences and other larger linguistic entities may also require `depend` attribute since they can be linked through relations such as cause and elaboration. |
| `equal` | Indicates a set *E* of elements which is coreferent with the current Linguistic element *C* (optional). Namely, the referent of *C* is equal to the referent of *E*, which is the referent of the element of *E* if *E* is a singleton set, and otherwise the set of the referents of the members of *E*. (The referent of a Linguistic element is represented by its self node.) *E* is a set of instances of the `LinguisticEntity DS`, the `Segment DS`, and the `Semantic DS`. |
| `semantics` | Indicates the semantic type of the plain-text part of the current Linguistic element (optional). Its value is a list of controlled terms. The default interpretation of this list is the sequential composition of the binary relations represented by the controlled terms, where a unary predicate *p* is regarded as a binary relation *r* such that *r*(*X*,*Y*) means that *p*(*X*) and *X*=*Y*. Non-default interpretations should be stipulated in the definitions of individual controlled terms. The self node of the current Linguistic element is the source (first) argument of this composed binary relation. |
| `compoundSemantics` | Indicates the semantic type of the current Linguistic element as a whole (optional). Its value is a list of controlled terms. The default interpretation of this list is the sequential composition of the binary relations represented by the controlled terms, where a unary predicate *p* is regarded as a binary relation *r* such that *r*(*X*,*Y*) means that *p*(*X*) and *X*=*Y*. Non-default interpretations should be stipulated in the definitions of individual controlled terms. The self node of the current Linguistic element is the source (first) argument of this composed binary relation. |
| `operator` | Indicates the semantic type of the function-word head (such as preposition, conjunction, and so on) of the current Linguistic element (optional). Its value is a list of controlled terms. The default interpretation of this list is the sequential composition of the binary relations represented by the controlled terms, where a unary predicate *p* is regarded as a binary relation *r* such that *r*(*X*,*Y*) means that *p*(*X*) and *X*=*Y*. Non-default interpretations should be stipulated in the definitions of individual controlled terms. The governor node and self node of the current Linguistic element are the source (first) and target (second) argument of this composed binary relation, respectively. When the function word is a coordinate conjunction, its meaning is the relationship among the conjuncts.<br><br>Just as with the `type` and `typelist` attributes in the `<Relation>` element, the `operator` attribute may partially specify the relation in question, in the sense that the exact relation is the composition *prq* of a prefix relation *p*, a postfix relation *q*, and the relation *r* specified by the attribute. *A p B* (*B q A*) means either *A* is equal to *B*, *B* is an instance, member, part or subset of *A*, or *B* is referred to by referring to *A* (metonymy). |
| `copy` | Specifies elements to copy prior to semantic interpretation (optional). The `id` values defined within the copied elements are substituted with new `id` values for the sake of the XML well-formedness. |
| `noCopy` | Specifies elements to exclude from copying due to the `copy` attribute. The |

| Name | Definition |
| --- | --- |
| | noCopy attribute in an element *C* refers to some descendant elements of the elements referenced by the copy attribute in *C*, and the copy operation excludes these descendant elements and their dependants. If the noCopy attribute refers to an ancestor element *E* of *C*, then *C* is replaced with a phrase coreferent with the copy of the minimal phrase containing *E*. (If the noCopy attribute points to multiple ancestors of *C*, then the smallest of them is *E*.) Otherwise *C* is replaced with the result of the copy. These operations are done prior to semantic interpretation (mapping to semantic structure). |
| substitute | Indicates the element to substitute with the current Linguistic element while copying due to the copy attribute of the parent Linguistic element (optional). The id values defined by the substituted element are also substituted with the corresponding id values defined by the current element. |
| inScope | Refers to the element (quantifier, modal operator, negation, abstraction, etc.) which introduces the minimal scope containing the referent of the current element. |
| edit | Indicates the original text string replaced by the annotator with the text in the element (optional). Its value is a colon (":") followed by this string. It begins with a colon to avoid empty value in the case of insertion. |

Semantics of LinguisticDocumentType:

| Name | Definition |
| --- | --- |
| LinguisticDocumentType | Describes an entire linguistic document, which can be a transcript, scenario, etc. The structure of a linguistic document is represented by a recursive hierarchy of linguistic entities: each linguistic entity in the document (section, paragraph, etc) can be broken down further into its component linguistic entities (other sections, sentences, syntactic constituents, etc.). LinguisticDocumentType extends LinguisticEntityType. |
| Heading | Describes a heading for a document or a division. |
| Division | Describes a textual division within a document, such as a chapter, a section, an embedded poem, etc. |
| Paragraph | Describes a paragraph within the document. |
| Sentences | Describes sentences occurring within the document. |
| Sentence | Describes a sentence: i.e., a phrase that is a complete proposition, question, request, etc., and does not participate in any syntactic dependency. Usually contains a period or a question mark at the end. |
| Quotation | Describes a direct narrative or citation. Namely, an utterance by someone other than the addressor of the surrounding linguistic content. |
| synthesis | Indicates the type of synthesis among the child elements and texts contained within the current element (optional). Default value for synthesis is coordination. |

Semantics of SentencesType:

| Name | Definition |
|---|---|
| SentencesType | Describes a sequence of sentences. SentencesType extends LinguisticEntityType. |
| Sentence | Describes a phrase that addresses a proposition, a question, a request, etc., and does not participate in any syntactic dependency (optional). Usually contains a period or a question mark at the end. |
| Quotation | Describes a direct narrative or citation (optional). |
| synthesis | Indicates the type of synthesis among the child entities and text contained within this entity (optional). Default value for synthesis is coordination. |

Semantics of SyntacticConstituentType:

| Name | Definition |
|---|---|
| SyntacticConstituentType | Describes a syntactic constituent. A syntactic constituent is a linguistic entity in a sentence that represents a semantic entity. For example, in the phrase "a big apple", the part "big apple" is a syntactic constituent but "a big" is not. SyntacticConstituentType extends LinguisticEntityType. |
| Head | Describes a syntactic constituent that may be the head or semantic representative of a larger constituent (optional). The Head element may or may not actually be a head linguistically, which can be useful for encoding ambiguity. |
| Phrase | Describes a syntactic constituent that is not the head of any larger constituent (optional). |
| Quotation | Describes a direct narrative or citation (optional). A Quotation element is phrasal, in the sense that it cannot be a head in a sentence. |
| baseForm | Indicates the base or uninflected form of the syntactic constituent (optional). |
| synthesis | Indicates the type of synthesis among the child entities and text of the current element (optional). The default value of synthesis is dependency. |
| functionWord | Indicates the function word or string of words that represent the operator (optional). For example, the function word of "on the beach" is "on", which gives a locative relation. The function word of "in order to escape" is "in order to", which represents the purpose. |

Semantics of synthesisType:

| Name | Definition |
|---|---|
| synthesisType | Indicates the type of synthesis that combines the child elements and texts of the current element. |
| none | Indicates that there is no direct semantic relation among child Linguistic elements and texts. |
| dependency | Indicates that the type of synthesis is dependency. Each child (Linguistic element or text) except one depends on a sibling Linguistic element or text. |
| forward | Indicates that the synthesis is a forward dependency chain. Each child linguistic entity except one depends on the closest sibling non-phrasal linguistic |
| | entity. The dependency should be forward (i.e. the governor should be to the right) if possible (that is, if there is a non-phrasal sibling linguistic entity to the right). The dependency relationships among the child linguistic entities are uniquely determined in a linguistic entity with an element-only content and `synthesis="forward"`. |
| `backward` | Indicates that the synthesis is a backward dependency chain. Each child linguistic entity except one depends on the nearest sibling non-phrasal linguistic entity. Contrary to `forward`, the dependency should be backward if possible. As with `forward`, the dependency relationships among the child linguistic entities are uniquely determined in a Linguistic element with an element-only content and `synthesis="backward"`. |
| `coordination` | Indicates that the synthesis is of the coordination type. Each child Linguistic element must be either a coordinate conjunction (such as "and" and "or") or an argument of it (coordinant). |
| `apposition` | Indicates that the synthesis is an apposition. When two linguistic entities form an apposition structure, the latter is a paraphrase, elaboration, etc. of the former. Each child Linguistic element of an element with `synthesis="apposition"` must be an apposition operator (such as "namely") or an argument of it. |
| `repair` | Indicates that the synthesis is a sequence of multiple linguistic entities where the last entity repairs the preceding erroneous ones. Each child Linguistic element of an element with `synthesis="repair"` must be a repair operator (such as "excuse me") or an argument of it. |
| `Error` | Indicates that the synthesis type is an error, which is the same as repair, except that all the child linguistic entities, including the last one, are erroneous. Each child Linguistic element of an element with `synthesis="repair"` must be a repair operator or an argument of it. |

## 4.4 Linguistic DS basic design (informative)

### 4.4.1 Introduction

The `Linguistic DS` describes the semantic structure of linguistic data independent of any particular language. Its essence is in a simple set of rules to map syntax (linguistic descriptions) to semantics, addressing potentially all the linguistic constructions across all the human natural languages. Of course it is not possible to detail how to describe all the specific sorts of linguistic constructions in all natural languages in the specification of a single description tool such as the `Linguistic DS`. However, it is possible to define how to describe linguistic data of any particular language, if the semantic structures of linguistic expressions in the language are agreed upon.

The `Linguistic DS` indirectly encodes the semantic structure of linguistic data using textual mark-up. That is, the semantic structure of the linguistic data is described by embedding or linking `Linguistic DS` tags to the textual linguistic data in such a way that the description can be mapped to the semantic structure. The `Linguistic DS` is designed so that its syntax and semantics can be related as directly as possible. The notions of dependency, syntactic constituent, and so on, are defined below so as to respect this guideline, though there are other, perhaps equally common, ways to define them. The semantic structures we assume below are compatible with many semantic representation languages such as those based on `Semantic DS` defined in ISO/IEC 15938-5.

The `Linguistic DS` specifies elements that describe linguistic entities. The elements form a hierarchy. At the lowest level of this hierarchy are syntactic constituents, such as words and phrases, occurring within sentences. At the higher levels are increasingly larger units of linguistic structures, such as sentences, paragraphs, other document divisions and entire documents. The `Linguistic DS` allows partial descriptions owing to the mixed content model, which simplifies the description by omitting tags. Not all linguistic entities need be Linguistic elements. Elements may have mixed contents in partial descriptions, whereas they have element-only or text-only contents in fuller descriptions.

There can be many ways to define what linguistic entities are. However, a semantics-oriented definition is adopted to respect the above guideline. Namely, a linguistic entity is a part of linguistic data representing a semantic entity. For instance, consider the sentence: “Tom and Mary live in a small house”. In this sentence, “Tom”, “Tom and Mary”, “in a small house”, “a small house”, “small house”, “house”, and so on, are linguistic entities (syntactic constituents when in sentences). Each of them represents a semantic entity as follows: “Tom and Mary” represents a group of two people, “small house” represents the notion of a small house, and so on. On the other hand, “Tom and”, “in a”, “a small”, and so forth are not linguistic entities because they fail to represent a integrated semantic unit like them.

Some linguistic accounts of agglutinative languages such as Japanese and Korean regard maximal morphological clusters (such as so-called *bunsetu* in Japanese) as syntactic constituents. However, the `Linguistic DS` does not necessarily do so. For instance, *tooi kuni* but not *kuni kara* (which is regarded as a *bunsetu*) is considered a syntactic constituent in the following Japanese expression:

```
 tooi    kuni    kara
distant country from
```
(“from distant countries”)

A linguistic entity corresponds to a semantic structure in a logical form which may be regarded as a network, or so-called frame representation, typed feature structure, etc. For instance, a syntactic constituent “for a boy” has the semantic structure encoded by the following logical form:

```
:r:beneficiary(G,S) & :u:boy(S) & :u:singular(S)
```

We refer to `G` and `S` here as the **governor node** and the **self node** of this semantic structure, respectively. Two semantic structures are combined to make a larger semantic structure by unifying the governor node of one of the two structures with the self node of the other. The above semantic structure is obtained from the following description.

```
<Phrase operator=":r:beneficiary" semantics=":u:boy :u:singular">
   for a boy
</Phrase>
```

In general, the `operator` attribute specifies the relation between the governor node and the self node of the semantic structure corresponding to the element. If the specified relation is a binary relation other than the identity, then the governor node and the self node are distinct. The `semantics` attribute describes the semantic content of the self node. More precisely speaking, both the `operator` attribute and the `semantics` attribute reflect the semantic structure of the plain-text children of the current element. So the following description is wrong, because “for” does not account for the value of the `semantics` attribute here.

```
<Phrase operator=":r:beneficiary" semantics=":u:boy :u:singular">
   for
   <Phrase>a boy </Phrase>
</Phrase>
```

The following description is fine.

```
<Phrase operator=":r:beneficiary">
   for
   <Phrase semantics=":u:boy :u:singular">a boy </Phrase>
</Phrase>
```

The `compoundSemantics` attribute encodes the semantic structure of the entire element. The following description is hence all right.

```
<Phrase compoundSemantics=":r:beneficiary :u:boy :u:singular">
   for
   <Phrase>a boy </Phrase>
</Phrase>
```

The following description tells that “pomme de terre” in French means a potato.

```
<Head xml:lang=”fr” compoundSemantics=":u:potato">
   pomme
   <Phrase>de
      <Phrase>terre </Phrase>
```

```
   </Phrase>
</Head>
```

Prefixes `:r:` and `:u:` indicate that the terms are binary relations and unary predicates, respectively; They are shortcuts for some URN prefixes defined beforehand in the description, for instance as follows:

```
<Header xsi:type="ClassificationSchemeAliasType" alias="r"
   href="urn:mpeg:mpeg7:cs:SemanticRelationCS:2001"/>

<Header xsi:type="ClassificationSchemeAliasType" alias="u"
   href="urn:SomeOntologyOfUnaryPredicates"/>
```

The above logical form may be regarded as a network as below:

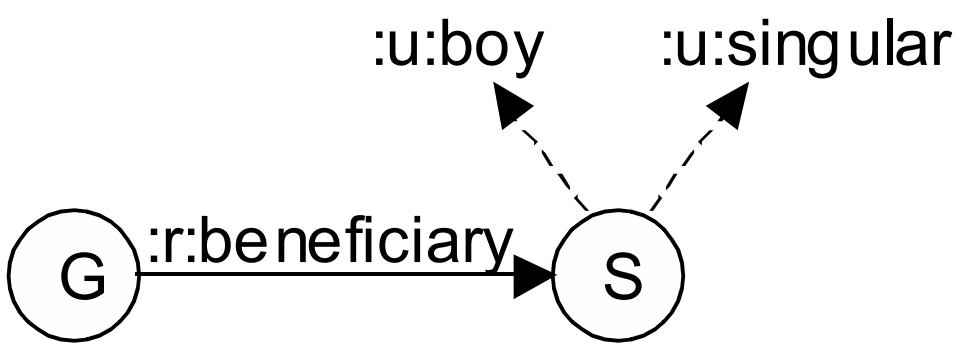

**Figure 1 The semantic structure of “for a boy”.**

Here an argument in the logical form is represented by a node in the network, a binary literal (an atomic formula with a binary relation) is represented by a solid arrow, labelled by the binary relation, from its first argument to its second argument, and a unary literal (an atomic formula with a unary relation) is represented by a broken arrow from its argument to the predicate node.

Precisely speaking, the `operator` attribute may partially specify the type of the relation between the governor node and the self node of the element, in the sense that some binary relation term may be missing at the beginning and/or end of the value of the `operator` attribute. The first argument of the relation specified by the `operator` attribute is either equal to, members of, instances of, parts of, subsets of, or indirectly (metonymicly) referenced by the governor node of the current element. Similarly, the second argument of the relation specified by the `operator` attribute is either equal to, members of, instances of, parts of, subsets of, or indirectly (metonymically) referenced by the self node of the current element.

For instance, consider the following example, which can involve metonymy.

```
<Mpeg7>
  <!-- MDS AMD/1 Example -->
  <Description xsi:type="ContentEntityType">
    <MultimediaContent xsi:type="LinguisticType">
      <Linguistic>
        <Sentence>
          Tom hit
          <Phrase operator="urn:SomeOntologyOfRelations:object">
            the apple pie
          </Phrase>.
        </Sentence>
      </Linguistic>
    </MultimediaContent>
  </Description>
</Mpeg7>
```

As mentioned previously, the `operator` attribute addresses the semantic relation between the governor and the current element. Provided that the above sentence involves a metonymy and means that Tom hit the person who ordered (made, liked, etc.) an apple pie, the `operator` attribute here partially describes the relation between “met” and “the apple pie” because the relation between the person and the apple pie is missing or implicit in this `operator` attribute. The following is a description with the complete value in the `operator` attribute.

```
<Mpeg7>
  <!-- MDS AMD/1 Example -->
  <Description xsi:type="ContentEntityType">
```

```
    <MultimediaContent xsi:type="LinguisticType">
      <Linguistic>
        <Sentence>
          Tom hit
          <Phrase operator="urn:SomeOntologyOfRelations:object
                            urn:SomeOntologyOfRelations:metonymy">
            the apple pie
          </Phrase>.
        </Sentence>
      </Linguistic>
    </MultimediaContent>
  </Description>
</Mpeg7>
```

Since it may be rather unrealistic to always require such a complete description concerning the `operator` attribute, however, the `Linguistic DS` allows partial descriptions such as the previous one.

See 4.4.5 for further account of this partiality of the `operator` attribute (and the `<Relation>` element), because it often makes sense in descriptions of cross references.

### 4.4.2 Dependency

Linguistic entities are recursively synthesized to form larger linguistic entities. Dependency is the most common way of this synthesis. The other types of synthesis are coordination, apposition, repair, and error, all of which are formally regarded as special cases of dependency in the `Linguistic DS`. There are many ways to define the notion of dependency, but we use the following definition to establish a direct mapping between syntax and semantics. That is, a syntactic constituent *X* depends on another syntactic constituent *Y* when the combination of *X* and Y represents a semantic entity that is equal to, a specialization of, or an instance of what *Y* represents. In "a small house", for instance, "a" depends on "small house" because "a small house" represents an instance of the concept of small houses. In "every man loves a woman", "every man" and "a woman" depend on "loves", because the sentence represents a state of affairs that is a specialization of the concept of loving. "Tom" depends on "for" in "for Tom" because "for Tom" represents a relationship with Tom that is a specialization of the binary relation represented by "for".

If a linguistic entity *X* depends on another linguistic entity *Y*, then we say *Y* **governs** *X* and call *Y* the governor of *X*. If in addition *X* and *Y* are next to each other, then their composition *Z* is a larger linguistic entity and *Y* is called the **head** of *Z.* In the case of "give me money", for instance, "give me" is the governor of "money" and the head of "give me money". "Give" is also a head of "give me money". A phrase is a syntactic constituent governing no neighbouring constituent. For instance, both "a" and "small" in "a small house" are phrases.

Let us write self(*X*) and gov(*X*) to denote the self node and the governor node, respectively, in the semantic structure corresponding to linguistic element *X*. (If *X* appears in an abstraction, then *X* may represent different semantic entities --- that is, self(*X*) may be different --- when referenced in different contexts. See the last paragraph of 4.4.5 for details.) If *X* depends on *Y* or *Y* is the head of *X*, then gov(*X*)=self(*Y*). For instance, "for" and "a boy" correspond to the semantic structures below:

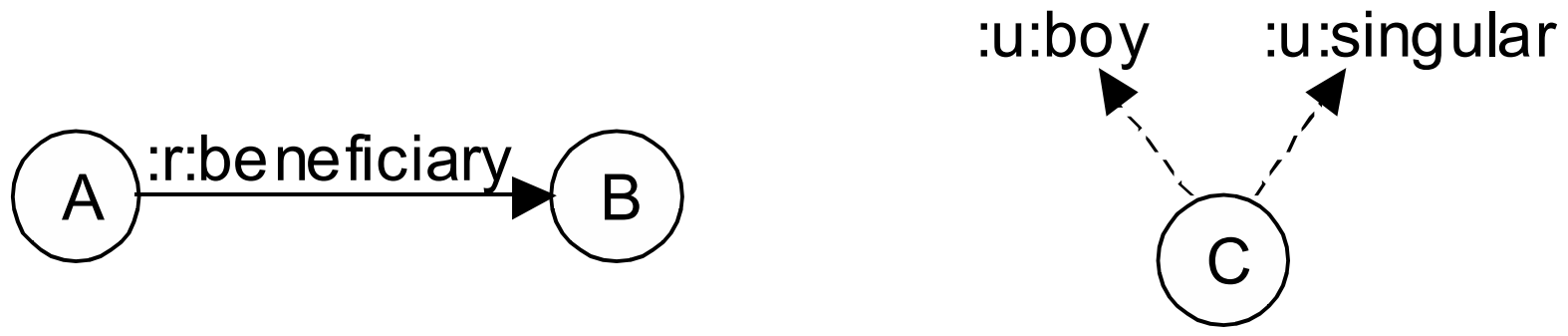

**Figure 2 The semantic structures of "for" (left) and "a boy" (right).**

Here self("for")=B, gov("for")=A, and self("a boy")=gov("a boy")=C hold. So we obtain the semantic structure in Figure 1 by letting gov("a boy")=self("for"), which is C=B. Of course the semantic structure of "a boy" is obtained by composing those of "a" and "boy", where self("a")=gov("a")=self("boy")=gov("boy")=C and "a" depends on "boy".

The following, more detailed description of "for a boy" encodes these dependencies.

```
<Phrase synthesis="dependency">
   <Head operator=":r:beneficiary">for </Head>
   <Phrase synthesis="dependency">
```

```
      <Phrase semantics=":u:singular">a </Phrase>
      <Head semantics=":u:boy">boy </Head>
   </Phrase>
</Phrase>
```

Here `synthesis=”dependency”` means that the element is synthesized by dependency: that is, each child element except the head of the whole element depends on another child element. The `Head` elements can (but need not) be governors of other elements, whereas the `Phrase` elements cannot. So the above description means that “a” depends on “boy” and “a boy” depends on “for”. We may omit `synthesis=”dependency”` because `dependency` is the default value for `synthesis` in a syntactic constituent. So a simplest description equivalent to the above one is:

```
<Phrase operator=":r:beneficiary">
  for
  <Phrase semantics=":u:boy">
     <Phrase semantics=":u:singular">a </Phrase>
     boy
  </Phrase>
</Phrase>
```

### 4.4.3 Coordination

Coordination is a second major way of synthesizing linguistic entities. “Tom and Mary”, “Tom or May”, “not only Tom but also Mary”, “Tom loves Mary and Bill loves Sue”, and so forth are coordinate structures. Coordination may be regarded as a special case of dependency, in which the coordinate conjunction is the head of the whole coordinate structure. For instance, “and” is the head of “Tom and Mary” and “or” is the head of “dead or alive”. Shown below is the semantic structure of “Tom and Mary”, where both the self node and the governor node are A, and :v: means that the operator (`:v:and`) is a variable-arity relation.

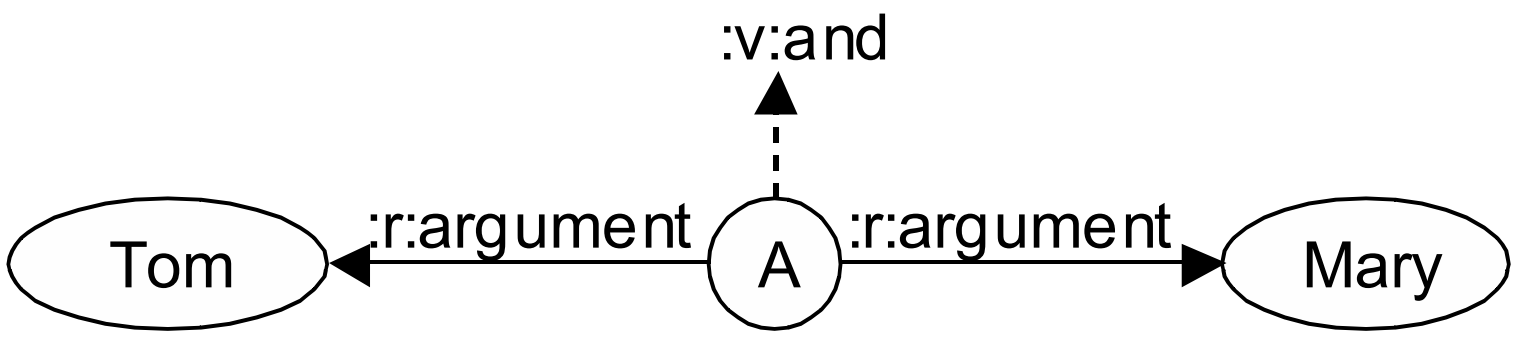

**Figure 3 The semantic structure of “Tom and Mary”.**

A description of “Tom and Mary” entailing this semantic structure is:

```
<Mpeg7>
   <!-- MDS AMD/1 Example -->
   <Description xsi:type="ContentEntityType">
      <MultimediaContent xsi:type="LinguisticType">
         <Linguistic>
            <Header xsi:type="ClassificationSchemeAliasType” alias="r"
               href="urn:mpeg:mpeg7:cs:SemanticRelationCS:2001"/>
            <Header xsi:type="ClassificationSchemeAliasType” alias="v"
               href="urn:SomeOntologyOfVariableArityRelations"/>
            <Sentence>
               <Phrase>
                  <Phrase operator=":r:argument">Tom </Phrase>
                  <Head operator=":v:and">and </Head>
                  <Phrase operator=":r:argument">Mary </Phrase>
               </Phrase>
               <!-- insert more elements -->
            </Sentence>
         </Linguistic>
      </MultimediaContent>
   </Description>
</Mpeg7>
```

`operator=”:r:argument”` may be omitted because `:v:and` presupposes the `:r:argument` relation for each of its arguments. The absence of the `operator` or `semantics` attribute has no default interpretation that its value is null. So a more usual description of “Tom and Mary” will be:

```
<Mpeg7>
   <!-- MDS AMD/1 Example -->
   <Description xsi:type="ContentEntityType">
      <MultimediaContent xsi:type="LinguisticType">
         <Linguistic>
            <Header xsi:type="ClassificationSchemeAliasType” alias="v"
               href="urn:SomeOntologyOfVariableArityRelations"/>
            <Sentence>
               <Phrase>
                  <Phrase>Tom </Phrase>
                  <Head operator=":v:and">and </Head>
                  <Phrase>Mary </Phrase>
               </Phrase>
               <!-- insert more elements -->
            </Sentence>
         </Linguistic>
      </MultimediaContent>
   </Description>
</Mpeg7>
```

or

```
<Mpeg7>
   <!-- MDS AMD/1 Example -->
   <Description xsi:type="ContentEntityType">
      <MultimediaContent xsi:type="LinguisticType">
         <Linguistic>
            <Header xsi:type="ClassificationSchemeAliasType” alias="v"
               href="urn:SomeOntologyOfVariableArityRelations"/>
            <Sentence>
               <Phrase operator=":v:and">
                  <Phrase>Tom </Phrase>
                  and
                  <Phrase>Mary </Phrase>
               </Phrase>
               <!-- insert more elements -->
            </Sentence>
         </Linguistic>
      </MultimediaContent>
   </Description>
</Mpeg7>
```

where `:v:and` is contributed by “and” in both cases, because the semantics and operator attributes describe the semantic content of the plain-text parts of the element.

Regarding “not only ... but also” as a compound coordination operator, “not only Tom but also Mary” may be described as follows, which means that `:v:notOnlyButAlso` is contributed by “not only” plus “but also”.

```
<Mpeg7>
   <!-- MDS AMD/1 Example -->
   <Description xsi:type="ContentEntityType">
      <MultimediaContent xsi:type="LinguisticType">
         <Linguistic>
            <Header xsi:type="ClassificationSchemeAliasType” alias="v"
               href="urn:SomeOntologyOfVariableArityRelations"/>
            <Sentence>
               <Phrase operator=":v:notOnlyButAlso">
                  not only
                  <Phrase>Tom </Phrase>
                  but also
                  <Phrase>Mary </Phrase>
```

```
                </Phrase>
                <!-- insert more elements -->
            </Sentence>
        </Linguistic>
    </MultimediaContent>
  </Description>
</Mpeg7>
```

See the subclause on syntax and semantics for apposition, repair, and error.

### 4.4.4 Ellipsis, abstraction and instantiation

Coordinate structures and repair structures are sometimes elliptical. Consider the following sentences:

*Tom loves Mary and Bill, Sue.*

*I gave this to Tom and that to Mary.*

*Tom wants to date with Mary, and Bill, Sue.*

In these examples, the second conjuncts ("Bill, Sue" and "that to Mary") lack their heads ("loves" and "gave", respectively). The `copy` and `substitute` attributes are used to describe these structures. Below is a description of "Tom wants to date with Mary, and Bill, Sue."

```
<Mpeg7>
  <!-- MDS AMD/1 Example -->
  <Description xsi:type="ContentEntityType">
    <MultimediaContent xsi:type="LinguisticType">
      <Linguistic>
        <Sentence synthesis="coordination">
          <Phrase id="TM">
            <Phrase id="TOM">Tom </Phrase>
            wants
            <Phrase>
              to
              <Phrase>
                date
                <Phrase>
                  with
                  <Phrase id="MARY">Mary</Phrase>
                </Phrase>
              </Phrase>
            </Phrase>
          </Phrase>,
          and
          <Phrase copy="#TM">
            <Phrase substitute="#TOM">Bill</Phrase>,
            <Phrase substitute="#MARY">Sue </Phrase>
          </Phrase>.
        </Sentence>
      </Linguistic>
    </MultimediaContent>
  </Description>
</Mpeg7>
```

The `copy` attribute means that "Bill, Sue" references "Tom wants to date with Mary" as a sort of lambda abstraction, and the `substitute` attribute specifies beta substitution. That is, the latter part of this coordinate structure is interpreted by copying "Tom wants to date with Mary" (together with the structural description, of course) while substituting "Tom" with "Bill" and "Mary" with "Sue", which results in "Bill wants to date with Sue". So the above description is semantically equivalent to the following:

```
<Mpeg7>
  <!-- MDS AMD/1 Example -->
  <Description xsi:type="ContentEntityType">
```

```
        <MultimediaContent xsi:type="LinguisticType">
            <Linguistic>
                <Sentence>
                    <Phrase>
                        <Phrase>Tom </Phrase>
                        wants
                        <Phrase>
                            to
                            <Phrase>
                                date
                                <Phrase>
                                    with
                                    <Phrase>Mary </Phrase>
                                </Phrase>
                            </Phrase>
                        </Phrase>
                    </Phrase>
                    and
                    <Phrase>
                        <Phrase>Bill </Phrase>
                        wants
                        <Phrase>
                            to
                            <Phrase>
                                date
                                <Phrase>
                                    with
                                    <Phrase>Sue</Phrase>
                                </Phrase>
                            </Phrase>
                        </Phrase>
                    </Phrase>.
                </Sentence>
            </Linguistic>
        </MultimediaContent>
    </Description>
</Mpeg7>
```

When a Linguistic element has `synthesis=”coordination”`, `synthesis=”apposition”`, `synthesis=”repair”`, or `synthesis=”error”`, a `copy` attribute of a child element *X* pointing to another child element *Y* may be omitted, though of course this omission is automatically recoverable only if some child element of *X* has some `substitute` attribute pointing to some descendant of *Y*. So `id=”TM”` and `copy=”#TM”` are not necessary in the previous description.

Ellipses occur also in constructions other than coordination, apposition, and so on. Comparative constructions very often involve ellipses. For instance, “Tom loves Mary better than Sue” has one, as this sentence is semantically equivalent to “Tom loves Mary better than Tom loves Sue” or “Tom loves Mary better than Sue loves Mary.” A description to encode the former interpretation follows.

```
<Mpeg7>
    <!-- MDS AMD/1 Example -->
    <Description xsi:type="ContentEntityType">
        <MultimediaContent xsi:type="LinguisticType">
            <Linguistic>
                <Header xsi:type="ClassificationSchemeAliasType” alias="r"
                    href="urn:SomeOntologyOfBinaryRelations"/>
                <Header xsi:type="ClassificationSchemeAliasType” alias="u"
                    href="urn:SomeOntologyOfUnaryPredicates"/>
                <Sentence>
                    <Head id="TomMary">
                        <Phrase>Tom </Phrase>
                        <Head>loves </Head>
                        <Phrase id="MARY">Mary </Phrase>
                    </Head>
                    <Phrase>
```

```
                      <Head semantics=":u:good" operator=":r:attribute">
                        better
                      </Head>
                      <Phrase>
                        <Head operator=":r:comparison">than </Head>
                        <Phrase copy="#TomMary">
                          <Phrase substitute="#MARY">Sue </Phrase>
                        </Phrase>
                      </Phrase>
                    </Phrase>
               </Sentence>
            </Linguistic>
        </MultimediaContent>
    </Description>
</Mpeg7>
```

The following, more complex, example entails that Tom lives in a house expected to be bigger than the house Mary lives in.

```
<Mpeg7>
    <!-- MDS AMD/1 Example -->
    <Description xsi:type="ContentEntityType">
        <MultimediaContent xsi:type="LinguisticType">
            <Linguistic>
                <Sentence id="TomLivesInAHouse">
                  <Phrase id="TOM">Tom </Phrase>
                  lives in a house
                  <Phrase id="EXPECT">
                    expected to be bigger than
                    <Phrase copy="#TomLivesInAHouse" noCopy="#EXPECT">
                      <Phrase substitute="#TOM">Mary </Phrase>
                      does
                    </Phrase>
                  </Phrase>.
                </Sentence>
            </Linguistic>
        </MultimediaContent>
    </Description>
</Mpeg7>
```

In general, the `noCopy` attribute refers to some descendant elements of the element referred to by the sibling `copy` attribute and excludes these descendant elements from the copy operation. So the above description means the following.

> The whole sentence (referred to as `TomLivesInAHouse`) minus “expected to be bigger than Mary does” is copied while substituting “Tom” with “Mary”, which results in “Mary lives in a house”.
>
> The phrase (“Mary does”) with the `noCopy` attribute in the original sentence is replaced with a phrase coreferent with the copy of the minimal phrase (“a house expected ...”) containing what the `noCopy` attribute points to (“expected to be bigger than Mary does”), which results in “Tom lives in a house bigger than *X*” where *X* corefers with “a house” in “Mary lives in a house.” (The fact that “a house ...” is the minmal phrase containing “expected to be bigger than Mary does” is not made explicit by the tags but can be automatically recognized.)

The two sentences “Tom lives in a house bigger than *X*” and “Mary lives in a house” do not stand in any left-to-right ordering; they belong to two separate text streams and none of them is left or right to the other.

A sometimes simpler way to fill an ellipsis is to use the `edit` attribute, which indicates that the text following the initial colon in its value is in the original data replaced with the content text in the element. A colon (“:”) is used to avoid an empty value for the attribute when describing an insertion. The following example replaces “,“ with “ loves” in “Tom loves Mary and Bill, Sue.”

```
<Mpeg7>
    <!-- MDS AMD/1 Example -->
    <Description xsi:type="ContentEntityType">
        <MultimediaContent xsi:type="LinguisticType">
```

```
         <Linguistic>
            <Sentence synthesis="coordination”>
               <Phrase>
                  <Phrase>Tom </Phrase>
                  loves
                  <Phrase>Mary </Phrase>
               </Phrase>
               and
               <Phrase>
                  <Phrase>Bill</Phrase>
                  <Head edit=":,"> loves</Head>
                  <Phrase>Sue </Phrase>
               </Phrase>.
            </Sentence>
         </Linguistic>
      </MultimediaContent>
   </Description>
</Mpeg7>
```

Another way to address the same interpretation is to use the `semantics` attribute as follows.

```
<Mpeg7>
   <!-- MDS AMD/1 Example -->
   <Description xsi:type="ContentEntityType">
      <MultimediaContent xsi:type="LinguisticType">
         <Linguistic>
            <Sentence>
               <Phrase>
                  <Phrase>Tom </Phrase>
                  loves
                  <Phrase>Mary </Phrase>
               </Phrase>
               <Head>and</Head>
               <Phrase semantics=”urn:someOntology:love”>
                  <Phrase>Bill</Phrase>,
                  <Phrase>Sue </Phrase>
               </Phrase>.
            </Sentence>
         </Linguistic>
      </MultimediaContent>
   </Description>
</Mpeg7>
```

Since the `semantics` attribute describes the plain-text part of the element, the above description captures the intended meaning, provided that “Bill” and “Sue” can be automatically recognized as the subject and the object of the elliptical verb, respectively.

The `inScope` attribute describes the scopes of abstractions such as quantifiers (*every*, *most*, *three*, etc.), modal operators (*may*, *probably*, *should*, etc.), and negations (*not*, *never*, etc.). Consider the following example of quantification.

```
<Mpeg7>
   <!-- MDS AMD/1 Example -->
   <Description xsi:type="ContentEntityType">
      <MultimediaContent xsi:type="LinguisticType">
         <Linguistic>
            <Sentence>
              <Phrase id="EVERY">Every </Phrase>man
              loves
              <Phrase inScope="#EVERY">a woman</Phrase>.
            </Sentence>
         </Linguistic>
      </MultimediaContent>
   </Description>
</Mpeg7>
```

This means that the referent of “a woman” is in the scope (more precisely, the body scope) of “every”, which entails that different men may love different women. The following example, on the other hand, means that all the men love one and the same woman, where `:d:top` means the top level of the discourse.

```
<Mpeg7>
   <!-- MDS AMD/1 Example -->
   <Description xsi:type="ContentEntityType">
      <MultimediaContent xsi:type="LinguisticType">
         <Linguistic>
            <Header xsi:type="ClassificationSchemeAlias” alias="d"
               href="urn:mpeg:mpeg7:cs:DeixisCS:2002"/>
            <Sentence>
              <Phrase>Every </Phrase>man
              loves
              <Phrase inScope=":d:top">a woman</Phrase>.
            </Sentence>
         </Linguistic>
      </MultimediaContent>
   </Description>
</Mpeg7>
```

The next example entails that Bill loves Tom’s wife (so-called the strict identity), because “his wife” is out of the scope of the abstraction of the first sentence and thus preserves the referent through the copy operation.

```
<Mpeg7>
   <!-- MDS AMD/1 Example -->
   <Description xsi:type="ContentEntityType">
      <MultimediaContent xsi:type="LinguisticType">
         <Linguistic>
            <Header xsi:type="ClassificationSchemeAlias” alias="d"
               href="urn:mpeg:mpeg7:cs:DeixisCS:2002"/>
            <Sentence id="TomLovesHisWife">
              <Phrase id="TOM">Tom </Phrase>
              loves
              <Phrase inScope=":d:top">
                 <Phrase equal=”#TOM”>his </Phrase>
                 wife
              </Phrase>.
            </Sentence>
            <Sentence copy="#TomLovesHisWife">
              So does
              <Phrase substitute="#TOM">Bill</Phrase>.
            </Sentence>
         </Linguistic>
      </MultimediaContent>
   </Description>
</Mpeg7>
```

On the other hand, the following entails that Bill loves Bill’s wife (so called the sloppy identity), as “his wife” here is in the scope of the abstraction and hence may have different referents across the copy.

```
<Mpeg7>
   <!-- MDS AMD/1 Example -->
   <Description xsi:type="ContentEntityType">
      <MultimediaContent xsi:type="LinguisticType">
         <Linguistic>
            <Sentence id="TomLovesHisWife">
              <Phrase id="TOM">Tom </Phrase>
              loves
              <Phrase inScope="#TomLovesHisWife">
                 <Phrase equal=”#TOM”>his </Phrase>
                 wife
              </Phrase>.
            </Sentence>
            <Sentence copy="#TomLovesHisWife">
```

```
            So does
            <Phrase substitute="#TOM">Bill</Phrase>.
         </Sentence>
      </Linguistic>
   </MultimediaContent>
</Description>
</Mpeg7>
```

### 4.4.5 Cross Reference

Cross references among linguistic entities other than through local syntheses are most often described by the `equal` attribute.

```
<Mpeg7>
   <!-- MDS AMD/1 Example -->
   <Description xsi:type="ContentEntityType">
      <MultimediaContent xsi:type="LinguisticType">
         <Linguistic>
            <Sentence>
               <Phrase id="J">John </Phrase>
               <Head>loves </Head>
               <Phrase>
                  <Phrase equal="#J" operator=":u:posessor">his </Phrase>
                  <Head>car </Head>
               </Phrase>.
            </Sentence>
         </Linguistic>
      </MultimediaContent>
   </Description>
</Mpeg7>
```

The `equal` attribute usually means that the self node of the current element is equal to the self node of the referenced element. So the above description entails that John possesses the car in question.

A more general sort of cross references are encoded by `Relation` elements, as usual in descriptions other than linguistic ones, too.

```
<Mpeg7>
   <!-- MDS AMD/1 Example -->
   <Description xsi:type="ContentEntityType">
      <MultimediaContent xsi:type="LinguisticType">
         <Linguistic>
            <Header xsi:type="ClassificationSchemeAliasType" alias="r"
               href="urn:mpeg:mpeg7:cs:SemanticRelationCS:2001"/>
            <Header xsi:type="ClassificationSchemeAliasType" alias="u"
               href="urn:SomeOntologyOfUnaryPredicates"/>
            <Sentence>
               <Phrase id="TOM" operator=":r:agent">Tom </Phrase>
               <Head semantics=":u:stop :u:past">turned </Head>
               <Phrase>
                  <Head operator=":r:direction">to </Head>
                  <Phrase>
                     <Phrase>the </Phrase>
                     <Head semantics=":u:right">
                        <Relation type=":r:argument" target="#TOM"/>
                        right
                     </Head>
                  </Phrase>
               </Phrase>.
            </Sentence>
         </Linguistic>
      </MultimediaContent>
   </Description>
</Mpeg7>
```

In general, a `Relation` element corresponds to a link from (the self node of) the source to (the self node of) the target. So for instance the above description has the following semantic structure.

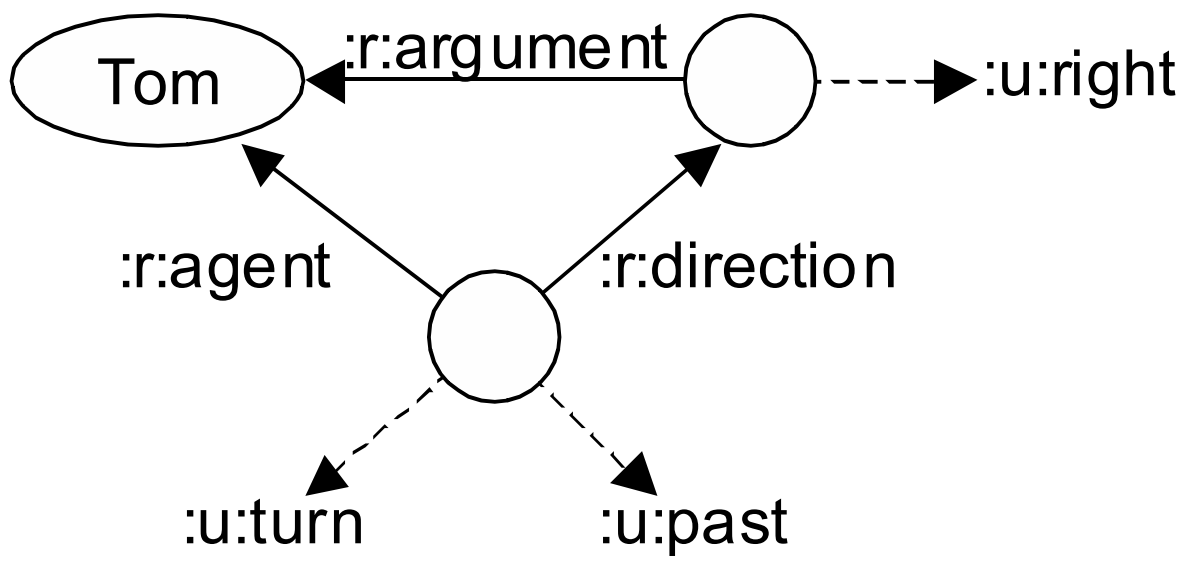

**Figure 4 The semantic structure of "Tom turned to the right.".**

The `Deixis CS` defines the deictic indices, which represent entities identifiable in the context at hand. The deictic indices can be the values of `equal`, `depend`, `source` and `target` attributes, such as follows, where deictic index `:d:p1` means the first person singular (I), and `:d:p2` the second person singular.

```
<Mpeg7>
   <!-- MDS AMD/1 Example -->
   <Description xsi:type="ContentEntityType">
      <MultimediaContent xsi:type="LinguisticType">
         <Linguistic>
            <Header xsi:type="ClassificationSchemeAliasType" alias="r"
               href="urn:mpeg:mpeg7:cs:SemanticRelationCS:2001"/>
            <Header xsi:type="ClassificationSchemeAliasType" alias="d"
               href="urn:mpeg:mpeg7:cs:DeixisCS:2002"/>
            <Header xsi:type="ClassificationSchemeAliasType" alias="d"
               href="urn:mpeg:mpeg7:cs:DeixisCS:2002"/>
            <Sentence>
               <Head>
                  <Relation type=":r:experiencer" generalTarget=":d:p1"/>
                  Hope
               </Head>
               to see
               <Phrase equal=":d:p2">you </Phrase>.
            </Sentence>
         </Linguistic>
      </MultimediaContent>
   </Description>
</Mpeg7>
```

Note that all these cross references pick up the meaning of the linguistic entities rather than the linguistic entity themselves. That is, the excerpt

```
<Relation type=":r:experiencer" generalTarget=":d:p1"/>
```

describes that the addressor ("I") is the experiencer of the hope but not of the linguistic entity "hope". To mention rather than use either argument of a relation, relation types defined in `SyntacticRelation CS`, `SyntacticSemanticRelation CS`, or `SemanticSyntacticRelation CS` are used. For instance, the following description means that John likes the name "John" but not John likes himself:

```
<Mpeg7>
   <!-- MDS AMD/1 Example -->
   <Description xsi:type="ContentEntityType">
      <MultimediaContent xsi:type="LinguisticType">
         <Linguistic>
            <Header xsi:type="ClassificationSchemeAliasType" alias="syn2"
               href="urn:mpeg:mpeg7:cs:SemanticSyntacticRelationCS:2002"/>
            <Sentence>
               <Phrase id="J">John </Phrase>
               likes
```

```
                <Phrase>
                    <Relation type=":syn2:equal" target="#J"/>
                    his name
                </Phrase>.
            </Sentence>
        </Linguistic>
    </MultimediaContent>
  </Description>
</Mpeg7>
```

The `type` and `typelist` attributes in the `<Relation>` element may partially specify the type of the relation between the source and the target arguments, in that some binary relation term may be missing at the beginning and/or end of the value of these attributes. The first argument of the relation specified by these attributes is either equal to, members of, instances of, parts of, subsets of, or indirectly (metonymically) referenced by the self node of the parent element. Similarly, the second argument of the relation specified by the `operator` attribute is either equal to, members of, instances of, parts of, subsets of, or indirectly (metonymically) referenced by the target argument specified by the `target` or `generalTarget` attribute. Also the `operator` attribute may partially specify the type of the relation between the governor node and the self node of the element in the same sense, as discussed in 4.1.

This partiality is useful typically in describing cross references. For instance, consider the following example.

```
<Mpeg7>
  <!-- MDS AMD/1 Example -->
  <Description xsi:type="ContentEntityType">
    <MultimediaContent xsi:type="LinguisticType">
        <Linguistic>
            <Sentence>
              <Phrase>Tom </Phrase>
              bought
              <Phrase id="CAR" operator=”urn:SomeOntologyOfRelations:object”>
                the car
                <Phrase>
                  <Relation type=”urn:SomeOntologyOfRelations:object”
                            target="#CAR"/>
                  designed by Bill
                </Phrase>
            </Sentence>.
        </Linguistic>
    </MultimediaContent>
  </Description>
</Mpeg7>
```

In the normal interpretation of this sentence, what Bill designed is a model of car and what Tom bought is an instance of it. So “the car designed by Bill” refers to this model of car, and the `<Relation>` element therein should refer to this model as the object of the designing event, whereas the object of the buying event is not exactly that model of car but an instance of it. The above description allows this interpretation, provided that the `operator` attribute may be partial, lacking the membership relation at the end of its value. Of course the same interpretation is possible also when the `operator` attribute is simply missing.

In the following example, on the other hand, not the `operator` attribute but the `type` attribute is partial, because what is sold here is not the content of the book, which Tom wrote, but some copies of the book.

```
<Mpeg7>
  <!-- MDS AMD/1 Example -->
  <Description xsi:type="ContentEntityType">
    <MultimediaContent xsi:type="LinguisticType">
        <Linguistic>
            <Sentence>
              <Phrase>Tom </Phrase>
              wrote
              <Phrase id="BOOK" operator=”urn:SomeOntologyOfRelations:object”>
                the book
                <Phrase>
                  <Relation type=”urn:SomeOntologyOfRelations:object”
```

```
                              target="#BOOK"/>
                   sold here
                 </Phrase>
             </Sentence>.
          </Linguistic>
       </MultimediaContent>
    </Description>
</Mpeg7>
```

In the following example, the subject “Every farmer who owns a donkey” is referred to twice, as the owner of a donkey and as the worker.

```
<Mpeg7>
  <!-- MDS AMD/1 Example -->
  <Description xsi:type="ContentEntityType">
    <MultimediaContent xsi:type="LinguisticType">
      <Linguistic>
        <Header xsi:type="ClassificationSchemeAliasType” alias="r"
          href="urn:mpeg:mpeg7:cs:SemanticRelationCS:2001"/>
        <Sentence>
          <Phrase operator=”:r:agent”>
            <Phrase>Every </Phrase>
            <Head id="FARMER">
              farmer
              <Phrase>
                <Phrase equal="#FARMER" operator=”:r:agent”>who </Phrase>
                owns
                <Phrase id=”DONKEY”>a donkey </Phrase>
              </Phrase>
            </Head>
          </Phrase>
          beats
          <Phrase equal=”#DONKEY”>it</Phrase>
        </Sentence>.
      </Linguistic>
    </MultimediaContent>
  </Description>
</Mpeg7>
```

Let *F* be the abstract type represented by (the self node of) “farmer who owns a donkey”. *F* is referenced (at least) twice in this sentence, and in both cases not the entire *F* but its instances (i.e., the non-specific individual owner and the non-specific individual worker) are used. So the two `operator` attributes (`operator=”:r:agent”`) lack a relation term (`urn:mpeg:mpeg7:cs:BaseRelationCS:2001:member` or something similar) at the end of their values. Also, a relation term (`urn:mpeg:mpeg7:cs:BaseRelationCS:2001:memberOf` or something similar) is missing at the beginning of the value of the first `operator=”:r:agent”`, because the referent of “work” (or the whole sentence) is an abstract type of working events whereas the first argument of the relation specified by this attribute should be a (non-specific) individual working event (or habit). The `Linguistic DS` allows the above partial description, because it is not realistic to always require complete descriptions (and because a complete value for the `operator` attribute can often be automatically inferred from the explicitly given partial value).

In general, if a Linguistic element *E* is contained in or equal to a Linguistic element *A* forming an abstraction and *E* represents a non-specific semantic entity *X* in the scope of that abstraction, then *E* when referenced within *A* represents *X*, whereas *E* when referenced from outside of *A* represents an instance of *X* outside the scope of the abstraction. In the above example, this holds for “farmer who owns a donkey” as *E* and also for “a donkey” as *E* (*A* is “farmer who owns a donkey” for both cases). Namely, “farmer who owns a donkey”, when referenced within itself, represents the non-specific farmer *f* in the scope of *F*, but when referenced from outside of itself, it represents a farmer $f_1$ who is an instance of *f* and outside the scope of *F*. Similarly, “a donkey” referenced within “farmer who owns a donkey” represents a non-specific donkey *d* in the scope of *F*, but when referenced from outside of “farmer who owns a donkey”, “a donkey” (hence “it” as well) represents an instance $d_1$ of *d* outside the scope of *F*. Although we omit futher technical details, $f_1$ owns $d_1$.

### 4.4.6 Multimedia Reference and Alignment

Cross-references between linguistic and other types of elements are also encoded by the `equal` attribute and the `Relation` element. The following example says that “this shot” is the video segment represented by the (possibly

VideoSegmentType) element referenced by http://i-content.org/movie.xml#scene4, and “the actress” is a part of the world represented by the shot.

```
<Mpeg7>
   <!-- MDS AMD/1 Example -->
   <Description xsi:type="ContentEntityType">
      <MultimediaContent xsi:type="LinguisticType">
         <Linguistic>
            <Sentence>
               <Phrase>
                 <Relation
                   type="urn:mpeg:mpeg7:cs:SemanticSyntacticRelationCS:2002:equal"
                   target="http://i-content.org/movie.xml#scene4"/>
                 the shot
               </Phrase>
               is very long.
            </Sentence>
           <Sentence>
               <Phrase>
                 <Relation
                   type="urn:mpeg:mpeg7:cs:SemanticRelationCS:2001:partOf"
                   target="http://i-content.org/movie.xml#scene4"/>
                 The actress
               </Phrase>
               is nice.
            </Sentence>
         </Linguistic>
      </MultimediaContent>
   </Description>
</Mpeg7>
```

SyntacticRelation CS, SyntacticSemanticRelation CS, and SemanticSyntacticRelation CS are again used to mention (as opposed to use) some arguments of relations. Those classification schemes apply not only to Linguistic elements but also to any other DS instances. For instance, the above example uses SemanticSyntacticRelation CS to mention the video shot rather than using its meaning (the world represented by the shot).

Elements of SpokenContentLatticeType, VideoTextType, ImageTextType, and TextSegmentType are interpreted as linguistic entities when they are referenced by the equal attribute and the Relation element. That is, if a Linguistic element refers to such an element by an equal attribute, the two elements represent the same semantic entity. If a Linguistic element *X* refers to such an element *Y* by a Relation element, then *X* and *Y* represent two semantic entities whose relation is specified by the type attribute of the Relation element. Of course, an element of SpokenContentLatticeType, VideoTextType, ImageTextType, or TextSegmentType is regarded as representing linguistic expression/utterance (its meaning is disregarded here) when it is mentioned (rather than used) by a Relation element (for instance, referenced as a second argument by a Relation element with a relation term in SemanticSyntacticRelation CS).

The Linguistic DS can describe not only inline textual data but also separate data, which may be text, video, or audio. The alignment with such data can be addressed by MediaLocator elements, start and length attributes, and Relation elements. A MediaLocator element specifies a portion of a separate data file, as shown in the following example meaning that the part from the 120th byte to the 133th byte in file transcript.txt is a sentence.

```
<Mpeg7>
   <!-- MDS AMD/1 Example -->
   <Description xsi:type="ContentEntityType">
      <MultimediaContent xsi:type="LinguisticType">
         <Linguistic>
            <Sentence>
               <MediaLocator xsi:type="StreamLocatorType">
                  <MediaUri>http://i-content.org/GDA/transcript.txt</MediaUri>
                  <StreamSection unit="byte" start="120" length="14"/>
               </MediaLocator>
            </Sentence>
         </Linguistic>
```

```
        </MultimediaContent>
    </Description>
</Mpeg7>
```

Similar descriptions are possible for video and audio data, as shown in the following example describing that the part from 421/25 second to 453/25 second in the video data `file://conversation.mpg` is a sentence, probably in the sense that an utterance of a sentence took place during that part of the video.

```
<Mpeg7>
    <!-- MDS AMD/1 Example -->
    <Description xsi:type="ContentEntityType">
        <MultimediaContent xsi:type="LinguisticType">
            <Linguistic>
                <Sentence>
                    <MediaLocator xsi:type="TemporalSegmentLocatorType">
                        <MediaUri>file://conversation.mpg</MediaUri>
                        <MediaTime>
                            <MediaTimePoint>T00:00:00:421F25</MediaTimePoint>
                            <MediaDuration>PT32N25F</MediaDuration>
                        </MediaTime>
                    </MediaLocator>
                    Hello.
                </Sentence>
            </Linguistic>
        </MultimediaContent>
    </Description>
</Mpeg7>
```

Note that such `MediaLocator` elements describe only the spatiotemporal alignment with linguistic data in separate files. More general, possibly semantic, relationships with data other than Linguistic elements are addressed by the `equal` attribute and the `Relation` element as discussed above.

It is often cumbersome and redundant to have a `MediaLocator` element in every Linguistic element to be aligned with an external data. To avoid this, the starting point and the length of the current linguistic entity may be described by the `start` and `length` attributes, respectively. A `start` attribute specifies a point in the file specified by the nearest preceding `MediaUri` element which is a child of an ancestor to the current element. When a `start` or `length` attribute has an `nonNegativeInteger` value, it assumes the same unit used last in the above-mentioned `MediaLocator` element.

The following example describes a sentence that consists of a phrase from the 122nd byte to the 127th byte (which contains a smaller phrase from the 124th byte to the 127th byte) and a head from the 128th byte to the 130th byte in the text file specified by the `MediaUri` element. Note that the `MediaLocator` element in this example is the nearest preceding one which is a child of an ancestor element to the elements containing the start and length attributes here. So the start attributes points into the file specified by the `MediaUri` element, and the unit of these start and length attribute is one byte.

```
<Mpeg7>
    <!-- MDS AMD/1 Example -->
    <Description xsi:type="ContentEntityType">
        <MultimediaContent xsi:type="LinguisticType">
            <Linguistic>
                <Sentence>
                    <MediaLocator xsi:type="TemporalSegmentLocatorType">
                        <MediaUri>http://i-content.org/GDA/transcript.txt</MediaUri>
                        <BytePosition offset="120" length="14"/>
                    </MediaLocator>
                    <Phrase start="122" length="6">
                        <Phrase start="124" length="4"/>
                    </Phrase>
                    <Head start="128" length="3"/>
                </Sentence>
            </Linguistic>
        </MultimediaContent>
    </Description>
```

```
</Mpeg7>
```

The following example describes a sentence that consists of a phrase ”John“ from 421/25 second to 434/25 second and a head verb ‘came’ from 436/25 second to 453/25 second in the date located as “file://transcript.mpg”.

```
<Mpeg>
   <!-- MDS AMD/1 Example -->
   <Description xsi:type="ContentEntityType">
      <MultimediaContent xsi:type="LinguisticType">
         <Linguistic>
            <Sentence>
               <MediaLocator xsi:type="TemporalSegmentLocatorType">
                  <MediaUri>file://transcript.mpg</MediaUri>
                  <MediaTime>
                     <MediaTimePoint>T00:00:00:421F25</MediaTimePoint>
                  </MediaTime>
               </MediaLocator>
               <Phrase start="T00:00:00:421F25" length="13N25F">John </Phrase>
               <Head start="T00:00:00:436F25" length="17N25F">came </Head>.
            </Sentence>
         </Linguistic>
      </MultimediaContent>
   </Description>
</Mpeg7>
```

The following example describes a sentence that consists of a phrase ”John“ from 421/25 second to 434/25 second and a head verb ‘came’ from 436/25 second to 453/25 second in the file located as “file://transcript.mpg”.

```
<Mpeg7>
   <!-- MDS AMD/1 Example -->
   <Description xsi:type="ContentEntityType">
      <MultimediaContent xsi:type="LinguisticType">
         <Linguistic>
            <Sentence>
               <MediaLocator xsi:type="TemporalSegmentLocatorType">
                  <MediaUri>file://transcript.mpg</MediaUri>
                  <MediaTime>
                     <MediaRelIncrTimePoint
                        mediaTimeUnit="PT1N25F">421</MediaRelIncrTimePoint>
                  </MediaTime>
               </MediaLocator>
               <Phrase start="421" length="13">John </Phrase>
               <Head start="436" length="17">came </Head>.
            </Sentence>
         </Linguistic>
      </MultimediaContent>
   </Description>
</Mpeg7>
```

Note that the above `start` and `length` attribute share the same unit specified by the `mediaTimeUnit` attribute.

As discussed above, the instances of `SpokenContentLatticeType`, `VideoTextType`, `ImageTextType`, and `TextSegmentType` are regarded as Linguistic elements when they are mentioned. So the alignments with such elements are addressed by `Relation` elements. In the example below, the target argument is regarded as a sentence.

```
<Mpeg7>
   <!-- MDS AMD/1 Example -->
   <Description xsi:type="ContentEntityType">
      <MultimediaContent xsi:type="LinguisticType">
         <Linguistic>
            <Sentence>
               <Relation type="urn:mpeg:mpeg7:SyntacticRelationCS:equal"
                  target="#anExternalSegment"/>
               Stop it.
            </Sentence>
```

```
         </Linguistic>
      </MultimediaContent>
   </Description>
</Mpeg7>
```

### 4.4.7 Guideline for simple description

Corresponding to various linguistic theories, there may be several different descriptions based on the Linguistic DS to address the same semantic structure. For instance, consider the description of “a man who came” below.

```
<Mpeg7>
   <!-- MDS AMD/1 Example -->
   <Description xsi:type="ContentEntityType">
      <MultimediaContent xsi:type="LinguisticType">
         <Linguistic>
            <Header xsi:type="ClassificationSchemeAliasType” alias="r"
               href="urn:mpeg:mpeg7:cs:SemanticRelationCS:2001"/>
            <Sentence>
               <Phrase id="MAN">
                  a man
                  <Phrase>
                     <Head eq="#MAN" id="WHO">who </Head>
                     <Phrase>
                        <Relation type=":r:agent" target="#WHO"/>
                        came
                     </Phrase>
                  </Phrase>
               </Phrase>
               <!-- insert more elements -->
            </Sentence>
         </Linguistic>
      </MultimediaContent>
   </Description>
</Mpeg7>
```

This is a rather exact encoding of the most common linguistic analysis of a relative clause. The following simpler description captures the same semantic structure. More precisely speaking, both of these descriptions fail to uniquely determine a semantic structure, but they specify the same range of semantic structures.

```
<Mpeg7>
   <!-- MDS AMD/1 Example -->
   <Description xsi:type="ContentEntityType">
      <MultimediaContent xsi:type="LinguisticType">
         <Linguistic>
            <Sentence>
               <Phrase id="MAN">
                  a man
                  <Phrase>
                     <Phrase eq="#MAN">who </Phrase>
                     came
                  </Phrase>
               </Phrase>
               <!-- insert more elements -->
            </Sentence>
         </Linguistic>
      </MultimediaContent>
   </Description>
</Mpeg7>
```

This latter description is of course perfectly valid, though different from the most common practice of linguistic theorization. For practical purposes, simpler descriptions such as this are more recommendable as far as they address the right semantic structures.

## 4.5 Linguistic DS example (informative)

### 4.5.1 Basic attributes

The following example describes a use of the `type` attribute to indicate that the word “salt” is used as a verb, provided that `:pos:` specifies some ontology of parts of speech.

```
<Head type=":pos:verb">salt </Head>
```

The `depend` attribute is attached to an extraposed linguistic entity depending on another entity embedded in one of its siblings. For example, in the sentence: “Tom, I think that Mary hates.”, “Tom” depends on “hates”, which is the head of embedded clause “Mary hates”.

```
<Mpeg7>
   <!-- MDS AMD/1 Example -->
   <Description xsi:type="ContentEntityType">
      <MultimediaContent xsi:type="LinguisticType">
         <Linguistic>
            <Header xsi:type="ClassificationSchemeAliasType” alias="r"
               href="urn:mpeg:mpeg7:cs:SemanticRelationCS:2001"/>
            <Sentence synthesis="backward">
               <Phrase depend="#H" operator=":r:patient">Tom </Phrase>,
               <Phrase>I </Phrase>
               think that
               <Phrase>
                  <Phrase>Mary </Phrase>
                  <Head id="H">hates </Head>
               </Phrase>.
            </Sentence>
         </Linguistic>
      </MultimediaContent>
   </Description>
</Mpeg7>
```

The following example indicates several operator types of  phrases

```
<Mpeg7>
   <!-- MDS AMD/1 Example -->
   <Description xsi:type="ContentEntityType">
      <MultimediaContent xsi:type="LinguisticType">
         <Linguistic>
            <Header xsi:type="ClassificationSchemeAliasType” alias="r"
               href="urn:mpeg:mpeg7:cs:SemanticRelationCS:2001"/>
            <Sentence>
               <Phrase operator=":r:experiencer">Tom </Phrase>
               <Head>loves </Head>
               <Phrase operator=":r:object">Sue </Phrase>.
            </Sentence>
            <Sentences synthesis="forward">
               <Sentence operator=":r:cause">Tom hit Mary. </Sentence>
               <Sentence>She cried. </Sentence>
            </Sentences>
         </Linguistic>
      </MultimediaContent>
   </Description>
</Mpeg7>
```

When the function word whose semantics is addressed by an `operator` attribute is a coordinate conjunction, its meaning is the relationship among the conjuncts, as in the following example. Note that the coordinate conjunction is the head of the coordinate structure.

```
<Mpeg7>
   <!-- MDS AMD/1 Example -->
   <Description xsi:type="ContentEntityType">
```

```
        <MultimediaContent xsi:type="LinguisticType">
            <Linguistic>
                <Header xsi:type="ClassificationSchemeAliasType” alias="v"
                    href="urn:SomeOntologyOfVariableArityRelations"/>
                <Sentence>
                    <Phrase>
                        <Phrase>Tom </Phrase>
                        <Head operator=":v:and">and </Head>
                        <Phrase>Mary </Phrase>
                    </Phrase>
                    got married.
                </Sentence>
            </Linguistic>
        </MultimediaContent>
    </Description>
</Mpeg7>
```

The following examples describe the base forms of the syntactic constituents in terms of the `baseForm` attribute.

```
<Head baseForm="grow">grew </Head>
<Head baseForm="dog">dogs </Head>
```

The following example involves an `edit` attribute indicating that the annotator replaced the comma in the original text with a white space followed by “loves.”

```
<Mpeg7>
    <!-- MDS AMD/1 Example -->
    <Description xsi:type="ContentEntityType">
        <MultimediaContent xsi:type="LinguisticType">
            <Linguistic>
                <Sentence>
                    <Phrase>Tom loves Mary </Phrase>
                    and
                    <Phrase>
                        Bill
                        <Head edit=":,">loves </Head>
                        Sue
                    </Phrase>
                </Sentence>
            </Linguistic>
        </MultimediaContent>
    </Description>
</Mpeg7>
```

### 4.5.2 Copy and substitution

The `id` values defined within the Linguistic element copied by a `copy` attribute are each substituted with new `id` values. For instance,

```
<a id="A"><b id="B"/><Relation target="#B"/></a>
<c id="C" copy="#A"/>
```

is regarded as equivalent with:

```
<a id="A"><b id="B"/><Relation target="#B"/></a>
<a id="C"><b id="B1"/><Relation target="#B1"/></a>
```

The `substitute` attribute indicates the element to substitute with the current Linguistic element while copying due to the `copy` attribute of the parent element. The `id` attributes and values defined by the substituted element are also substituted with the corresponding `id` attributes and values defined by the current element. For instance,

```
<a id="A"><b id="B"/><Relation target="#B"/></a>
<c id="C" copy="#A"><d substitute="#B"><e/></d></c>
```

is regarded as equivalent with:

```
<a id="A"><b id="B"/><Relation target="#B"/></a>
<a id="C"><d id="B1"><e/></d><Relation target="#B1"/></a>
```

The following example, with multiple ID values in the `copy` attribute, entails that Bill is both tall and heavy, because the `copy` attribute and the `substitute` attribute specify to copy “Tom is tall” and “He is heavy, too” while substituting `<Phrase id=”TOM”>Tom </Phrase>` with `<Phrase id=”BILL”>Bill</Phrase>` and `TOM` with `BILL` (in `equal=”#TOM”`).

```
<Mpeg7>
   <!-- MDS AMD/1 Example -->
   <Description xsi:type="ContentEntityType">
      <MultimediaContent xsi:type="LinguisticType">
         <Linguistic>
            <Sentence id="TALL"><Phrase id="TOM">Tom </Phrase>is tall. </Sentence>
            <Sentence id="HEAVY">
               <Phrase equal="#TOM">He </Phrase>is heavy, too.
            </Sentence>
            <Sentence copy="#TALL #HEAVY">
              So is <Phrase substitute="#TOM">Bill</Phrase>.
            </Sentence>
         </Linguistic>
      </MultimediaContent>
   </Description>
</Mpeg7>
```

### 4.5.3 Dependency and other syntheses

The `Head` element may or may not actually be a head, which can be useful for encoding ambiguity. The following example accommodates two interpretations: “planes which are quickly flying” and “to fly planes quickly”:

```
<Phrase>
   <Head>
      <Phrase>quickly</Phrase>
      flying
   </Head>
   <Head>planes </Head>
</Phrase>
```

The following two examples illustrate the use of `dependency` value for the `synthesis` attribute. The interpretation of the example below may be to fly planes (where “planes” depends on “flying”) or planes which are flying (where “flying” depends on “planes”), because the default value for `synthesis` for `Phrase` elements is `dependency`. Note that the head is not specified here, and is therefore left open:

```
<Phrase>flying planes </Phrase>
```

When the head is specified uniquely and explicitly, the dependency relationships among the children are uniquely determined as below, where both “the” and “good” depends on “idea”:

```
<Phrase>
   <Phrase>the </Phrase>
   <Phrase>good </Phrase>
   <Head>idea </Head>
</Phrase>
```

In the following example, “quickly” depends on “flying” and “flying” depends on “planes” due to `synthesis=”forward”`

```
<Phrase synthesis="forward">
   <Phrase>very </Phrase>
   <Head>quickly </Head>
   <Head>flying </Head>
   <Head>planes </Head>
```

```
</Phrase>
```

Note that using `forward` simplifies the description. In fact, the above description is much simpler than the following, equivalent one.

```
<Phrase>
   <Phrase>
      <Phrase>
         <Phrase>very </Phrase>
         <Head>quickly </Head>
      </Phrase>
      <Head>flying </Head>
   </Phrase>
   <Head>planes </Head>
</Phrase>
```

In the following example, “eat” depends on “to” and “to” depends on “want”, because “to” is the nearest potential head before “eat”, and “want” is the nearest potential head before “to”.

```
<Head synthesis="backward">
    <Head>want </Head>
    <Head>to </Head>
    <Phrase>eat </Phrase>
</Head>
```

As with the use of `forward`, `backward` simplifies the description:

```
<Head>
    <Head>want </Head>
    <Phrase>
        <Head>to </Head>
        <Phrase>
            <Head>eat </Head>
        </Phrase>
     </Phrase>
</Head>
```

Below is an example of apposition structure.

```
<Mpeg7>
   <!-- MDS AMD/1 Example -->
   <Description xsi:type="ContentEntityType">
      <MultimediaContent xsi:type="LinguisticType">
         <Linguistic>
            <Sentence>
               <Phrase>I </Phrase>
               <Head synthesis="apposition">
                  <Head>introduced Mary to Sue, </Head>
                  that is,
                  <Head>
                     <Head edit=":">introduced </Head>
                     my girlfriend to my wife
                  </Head>
               </Head>
            </Sentence>
         </Linguistic>
      </MultimediaContent>
   </Description>
</Mpeg7>
```

The `synthesis=”repair”` in the following example means that “gave Mary to the dog” is repaired by “(gave) the dog to Mary”.

```
<Mpeg7>
   <!-- MDS AMD/1 Example -->
```

```
    <Description xsi:type="ContentEntityType">
       <MultimediaContent xsi:type="LinguisticType">
          <Linguistic>
             <Sentence>I
                <Head synthesis="repair">
                   <Head>gave Mary to the dog, </Head>
                   oh, I'm sorry,
                   <Head>
                      <Head edit=":">gave </Head>
                      the dog to Mary
                   </Head>
                </Head>.
             </Sentence>
          </Linguistic>
       </MultimediaContent>
    </Description>
</Mpeg7>
```

### 4.5.4 Quotation

The following example shows how to describe a direct narrative by a `Quotation` element. Note that this description almost determines the syntactic structure of the whole sentence, because a `Quotation` element is phrasal and hence cannot govern other syntactic constituents.

```
<Mpeg7>
   <!-- MDS AMD/1 Example -->
   <Description xsi:type="ContentEntityType">
      <MultimediaContent xsi:type="LinguisticType">
         <Linguistic>
            <Header xsi:type="ClassificationSchemeAliasType" alias="r"
               href="urn:mpeg:mpeg7:cs:SemanticRelationCS:2001"/>
            <Sentence>
               <Quotation>I quit </Quotation>,
               <Head>said </Head>
               <Phrase operator=":r:agent">Sue </Phrase>.
            </Sentence>
         </Linguistic>
      </MultimediaContent>
   </Description>
</Mpeg7>
```

### 4.5.5 Partial description

The `Linguistic DS` allows partial description of linguistic structure. The following shows a description of the sentence: “You might want to suppose that flying planes may be dangerous.” The example specifies that “flying” depends on “planes” by specifying that the “flying” is a phrase and thus cannot be a governor. The relations among the child entity and pieces of child texts are left undescribed; here it is just assumed that some dependencies hold among them, without committing to any further details.

```
<Mpeg7>
   <!-- MDS AMD/1 Example -->
   <Description xsi:type="ContentEntityType">
      <MultimediaContent xsi:type="LinguisticType">
         <Linguistic>
            <Sentence>
               You might want to suppose that
               <Phrase>flying </Phrase>
               planes may be dangerous.
            </Sentence>
         </Linguistic>
```

```
        </MultimediaContent>
    </Description>
</Mpeg7>
```

On the other hand, the following example describes the syntactic structure for the same sentence in more detail. In this connection, the `synthesis` attribute addresses accurate characterizations of the type of combination among the child elements and texts, and thus simplifies the description.

```
<Mpeg7>
    <!-- MDS AMD/1 Example -->
    <Description xsi:type="ContentEntityType">
        <MultimediaContent xsi:type="LinguisticType">
            <Linguistic>
                <Sentence>
                    <Phrase>You</Phrase>
                    <Head>might</Head>
                    <Phrase>
                        <Head>want</Head>
                        <Phrase>
                            <Head>to</Head>
                            <Phrase>
                                <Head>suppose</Head>
                                <Phrase>
                                    <Head>that</Head>
                                    <Phrase>
                                        <Phrase>
                                            <Phrase>
                                                <Head>flying</Head>
                                            </Phrase>
                                            <Head>planes</Head>
                                        </Phrase>
                                        <Head>may be dangerous</Head>
                                    </Phrase>
                                </Phrase>
                            </Phrase>
                        </Phrase>
                    </Phrase>.
                </Sentence>
            </Linguistic>
        </MultimediaContent>
    </Description>
</Mpeg7>
```

### 4.5.6 Recommended granularity of description

The following example illustrates a rather simple (in terms of the number of tags and attributes) description that uniquely determines the prepositional content of the paragraph, in the practical sense that a reasonable automatic analysis of the paragraph under the constraint provided by this description will uniquely determine the syntactic structures, anaphora, and coreferences. So the example indicates a recommended granularity of linguistic annotation for the sake of practical applications such as information retrieval, summarization, translation, and so on, though the really minimal description would have much fewer tags.

The first sentence is “This is Akashi Channel Bridge, which connects Kobe City and Awaji Island of Hyogo Prefecture”, which contains a relative clause. The second and third sentences are “Look. It's so big. It's the world's longest suspension bridge, whose length is about 4,000 meters,” where the object of “look” is elliptical and is coreferent with “this” in the first sentence. The last is a cleft sentence “It's two wires which support the weight of the bridge, which is as much as 150,000 tons.”

```
<Mpeg7>
    <!-- MDS AMD/1 Example -->
    <Description xsi:type="ContentEntityType">
        <MultimediaContent xsi:type="LinguisticType">
```

```
        <Linguistic>
            <Paragraph>
                <Sentence>
                    <Phrase id="THIS">This </Phrase>
                    is
                    <Head id="ACB">Akashi Channel Bridge </Head>,
                    <Phrase synthesis="backward">
                        <Phrase equal="#ACB">which </Phrase>
                        connects
                        <Head>
                            <Phrase>Kobe City </Phrase>
                            and
                            <Phrase>Awaji Island </Phrase>
                        </Head>
                        of
                        <Phrase>Hyogo Prefecture </Phrase>
                    </Phrase>.
                </Sentence>
                <Sentence>
                    Look.
                    <Relation type=":r:object" target="#THIS"/>
                </Sentence>
                <Sentence>
                    <Phrase equal="#THIS">It</Phrase>'s
                    <Phrase>so big</Phrase>.
                </Sentence>
                <Sentence>
                    <Phrase equal="#ACB">It</Phrase>'s
                    <Head id="WLSB">
                        the
                        <Phrase>world's longest </Phrase>
                        <Phrase>suspension </Phrase>
                        bridge
                    </Head>,
                    <Phrase>
                        <Phrase>
                            <Phrase equal="#WLSB">whose </Phrase>
                            length
                        </Phrase>
                        is about
                        <Phrase>4,000 meters</Phrase>
                    </Phrase>.
                </Sentence>
                <Sentence>
                    <Phrase id="IT">It</Phrase>'s
                    <Phrase>two wires </Phrase>
                    <Phrase depend="#IT">
                        <Phrase equal="#IT">which </Phrase>
                        support
                        <Phrase id="W">
                            <Phrase>the </Phrase>
                            weight
                            <Phrase>
                                of
                                <Phrase equal="#ACB">
                                    the bridge
                                </Phrase>
                            </Phrase>,
                            <Phrase>
                                <Phrase equal="#W">which </Phrase>
                                is as much as
                                <Phrase>150,000 tons</Phrase>
                            </Phrase>
                        </Phrase>
                    </Phrase>.
```

```
                  </Sentence>
              </Paragraph>
          </Linguistic>
      </MultimediaContent>
  </Description>
</Mpeg7>
```

# 5 Classification Schemes

## 5.1 DeixisCS

```
<!-- ##################################################  -->
<!-- Definition of Deixis CS  (AMD/1)                    -->
<!-- ##################################################  -->
<ClassificationScheme uri="urn:mpeg:mpeg7:cs:DeixisCS:2002">
   <Term termID="p0">
      <Definition xml:lang="en">General public </Definition>
   </Term>
   <Term termID="p1">
      <Definition xml:lang="en">First person singular (`I') </Definition>
   </Term>
   <Term termID="p1p">
      <Definition xml:lang="en">First person plural (`We') </Definition>
   </Term>
   <Term termID="p1i">
      <Definition  xml:lang="en">First  person  plural  inclusive  (`We'  including
`you')</Definition>
   </Term>
   <Term termID="p1x">
      <Definition  xml:lang="en">First  person  plural  exclusive(`We'  excluding
`you')</Definition>
   </Term>
   <Term termID="p2">
      <Definition xml:lang="en">Second person singular (Singular `you')</Definition>
   </Term>
   <Term termID="p2p">
      <Definition xml:lang="en">Second person plural (Plural `you') </Definition>
   </Term>
   <Term termID="nil">
      <Definition xml:lang="en">Nothing </Definition>
   </Term>
   <Term termID="top">
      <Definition xml:lang="en">The top-level discourse </Definition>
   </Term>
   <Term termID="self">
      <Definition xml:lang="en">The element itself </Definition>
   </Term>
</ClassificationScheme>
```

## 5.2 SyntacticRelationCS

```
<!-- ##################################################  -->
<!-- Definition of SyntacticRelation CS (AMD/1)          -->
<!-- ##################################################  -->
<ClassificationScheme uri="urn:mpeg:mpeg7:cs:SyntacticRelationCS:2002">
   <Header xsi:type="DescriptionMetadataType">
      <Comment>
         <FreeTextAnnotation xml:lang="en">
   This set of relations relate linguistic entities themselves
```

```
   rather than their meanings. Namely, both the first and the
   second arguments are mentioned rather than used.
   </FreeTextAnnotation>
      </Comment>
   </Header>
   <Import href="urn:mpeg:mpeg7:cs:SemanticRelationCS:2001"/>
   <Term termID="null">
      <Definition xml:lang="en">Empty</Definition>
   </Term>
</ClassificationScheme>
```

### 5.3 SyntacticSemanticRelationCS

```
<!-- #####################################################  -->
<!-- Definition of SyntacticSemanticRelation CS      (AMD/1)       -->
<!-- #####################################################  -->
<ClassificationScheme uri="urn:mpeg:mpeg7:cs:SyntacticSemanticRelationCS:2002">
   <Header xsi:type="DescriptionMetadataType">
      <Comment>
         <FreeTextAnnotation xml:lang="en">
   This set of relations relate the first argument as
   the linguistic entity itself and the meaning of the
   second argument. That is, the first argument is mentioned
   but the second argument is used.
         </FreeTextAnnotation>
      </Comment>
   </Header>
   <Import href="urn:mpeg:mpeg7:cs:SemanticRelationCS:2001"/>
   <Term termID="null">
      <Definition xml:lang="en">Empty</Definition>
   </Term>
</ClassificationScheme>
```

### 5.4 SemanticSyntacticRelationCS

```
<!-- #####################################################  -->
<!-- Definition of SemanticSyntacticRelation CS    (AMD/1)               -->
<!-- #####################################################  -->
<ClassificationScheme uri="urn:mpeg:mpeg7:cs:SemanticSyntacticRelationCS:2002">
   <Header xsi:type="DescriptionMetadataType">
      <Comment>
         <FreeTextAnnotation xml:lang="en">
   This set of relations relate the meaning of the
   first argument and the second argument as the linguistic
   entity itself. That is, the first argument is used
   but the second argument is mentioned.
         </FreeTextAnnotation>
      </Comment>
   </Header>
   <Import href="urn:mpeg:mpeg7:cs:SemanticRelationCS:2001"/>
   <Term termID="null">
      <Definition xml:lang="en">Empty</Definition>
   </Term>
</ClassificationScheme>
```